\documentclass{clv3}

\usepackage{hyperref}
\usepackage{xcolor}
\definecolor{darkblue}{rgb}{0, 0, 0.5}
\hypersetup{colorlinks=true,citecolor=darkblue, linkcolor=darkblue, urlcolor=darkblue}
\usepackage{url,hyperref,lineno,microtype,subcaption}
\usepackage{lingmacros}
\usepackage{todonotes}
\usepackage{dsfont}

\issue{1}{1}{2016}

\runningtitle{Schr{\"o}dinger’s Tree}

\runningauthor{Kulmizev and Nivre}

\begin{document}

\title[Schr{\"o}dinger’s Tree]{Schr{\"o}dinger’s Tree -- On Syntax and Neural Language Models} 

\author{Artur Kulmizev\thanks{E-mail: \texttt{artur.kulmizev@lingfil.uu.se}}}
\affil{Uppsala University}

\author{Joakim Nivre}
\affil{Uppsala University}

\maketitle

\begin{abstract}
In the last half-decade, the field of natural language processing (NLP) has undergone two major transitions: the switch to neural networks as the primary modeling paradigm and the homogenization of the training regime (pre-train, then fine-tune). Amidst this process, language models have emerged as NLP's workhorse, displaying increasingly fluent generation capabilities and proving to be an indispensable means of knowledge transfer downstream. Due to the otherwise opaque, black-box nature of such models, researchers have employed aspects of linguistic theory in order to characterize their behavior. Questions central to syntax --- the study of the hierarchical structure of language --- have factored heavily into such work, shedding invaluable insights about models' inherent biases and their ability to make human-like generalizations. In this paper, we attempt to take stock of this growing body of literature. In doing so, we observe a lack of clarity across numerous dimensions, which influences the hypotheses that researchers form, as well as the conclusions they draw from their findings. To remedy this, we urge researchers make careful considerations when investigating coding properties, selecting representations, and evaluating via downstream tasks. Furthermore, we outline the implications of the different types of research questions exhibited in studies on syntax, as well as the inherent pitfalls of aggregate metrics. Ultimately, we hope that our discussion adds nuance to the prospect of studying language models and paves the way for a less monolithic perspective on syntax in this context.
\end{abstract}

\section{Introduction}
\label{sec:introduction}

Syntax --- how words are combined to form sentences in natural language --- has perhaps never garnered as much attention from NLP researchers as it does in the present day. Naturally, its recent relevance at conferences is owed to the deep learning paradigm, which the NLP community has embraced with open arms since the midpoint of the last decade. Prior to this paradigm shift, questions central to syntax 
were often restricted to the parsing domain. There, researchers were largely interested in developing supervised algorithms for processing structured input --- usually in the form of annotated constituency or dependency treebanks. Beyond parsing, syntax also often factored into researchers' hypotheses about what information models may need to succeed in a given task. Feature engineering was a pivotal component of pre-neural NLP, where text was filtered through hand-crafted feature templates that emphasized parts of speech, morphology, and tree structure, so as to inform simple, often linear models about the underlying syntax of sentences. 

The deep learning revolution of the mid 2010s quickly obviated the need for feature engineering, which was widely considered a time-consuming and painstaking process. Embeddings --- dense vectors representing the distributional properties of words --- quickly replaced the sparse, hand-crafted vectors of yore and boosted performance dramatically \citep{mikolov13,pennington14}. Such progress presented a trade-off, however: accuracy at the expense of interpretability. Indeed, without the guiding hand of the feature engineer, it became difficult to ascertain what properties of natural language the new neural models --- highly complex and non-linear --- had come to rely on. 

It was this uncertainty that inspired a new line of inquiry within NLP, concerning what exactly models know and how they come to learn it. Early insights from this domain intimated that neural networks could capture facets of the hierarchical structure of language, beyond the linear order of words in a sentence. The Long Short Term Memory network (LSTM) \citep{hochreiter97} featured prominently in such studies, where researchers employed linguistic 
minimal pairs (mostly based on agreement phenomena) in order to demonstrate the model's sensitivity to syntactic hierarchy \citep{linzen2016assessing,gulordava2018colorless}. Such findings were deemed exciting mainly due the LSTM's design as a sequence processor, which lacked the sort of structural supervision or inductive bias that one might encounter in the parsing literature. 

Amidst skyrocketing research budgets and the continued advancement of processing hardware, NLP faced another paradigm shift in 2018--19. Researchers began realizing that representations for input words need not be fixed to a single static vector per type (as with word embeddings), but can instead be computed dynamically, with each word contextualized with respect to the rest of the sentence \citep{peters18}. Per this logic, it also became apparent that models capable of generating such representations could be fine-tuned with respect to downstream tasks, with impressive gains in performance thereafter \citep{howard-ruder-2018-universal}. Language models --- the basis of classic word embedding algorithms --- were a natural fit for this paradigm and became NLP's backbone going forward. 

In the modern day, models like BERT \citep{devlin19}, GPT \citep{radford2019language} and their successors feature prominently in NLP research, showcasing the efficacy of the pretrain-and-finetune paradigm. Naturally, the human-like generation capability of such models, as well as their success on natural language understanding (NLU) benchmarks \citep{wang2018glue,wang2019superglue}, makes the question of what the models know about language and how they acquire such knowledge and ever-pressing one. Increasingly, we find, NLP researchers turn to the field of syntax --- with its decades of research, theory, and debate --- in order to answer such questions. 
In this paper, we attempt to take stock of the ever-growing literature on the syntactic capabilities of neural language models. In doing so, we observe a lack of clarity across numerous dimensions, which influences the hypotheses that researchers form, as well as the conclusions they draw from their findings. We argue that this failure of articulation results in a body of work whose hypotheses, methodologies, and conclusions comprise many conflicting insights, giving rise to a paradoxical picture reminiscent of Schr\"{o}dinger's cat --- where syntax appears to be simultaneously dead and alive inside the black box models. In particular, by framing studies around aggregate metrics and benchmarks, syntax is often reduced to a monolithic phenomenon, which fails to do justice both to the complex interplay between different manifestations of hierarchical structure in natural language and to the substantial variation that exists across typologically different languages.

Our goal in this article is not to criticize earlier studies, which all provide valuable pieces of 
evidence for understanding the role of syntax in contemporary NLP, particularly language models. Instead, we propose a number of conceptual distinctions, the consideration and articulation of which, we argue, can help us better understand the seemingly conflicting results, resolve some of the apparent contradictions, and pave the way for a more nuanced and articulated research agenda. 
To provide the necessary background for this analysis, we begin by introducing the concept of syntax from a bird's eye perspective. We then review a representative sample of investigations into the syntactic capabilities of neural language models, which we categorize as belonging to three different paradigms. We supplement this review by discussing what we perceive to be important distinctions about syntax left implicit in this body of work. This leads to a discussion of different classes of research questions underlying the surveyed literature, and the role of aggregate metrics in addressing these research questions. We conclude with some thoughts on how our analysis can inform our research methodology for the future.

\section{Background: Aspects of Syntax} \label{aspects}
\label{sec:background}
Syntax is usually described as the way that words are combined into larger expressions like phrases and sentences. On one hand, syntax can then be contrasted with morphology, which is concerned with the internal structure of words. On the other hand, it can be contrasted with semantics, which deals with the \emph{meaning} of words, phrases and sentences --- as opposed to their \emph{form}. In reality, however, syntax is concerned with the complex mapping between \emph{form} and \emph{meaning} at the phrase and sentence level. It is therefore important to make a distinction between \emph{syntactic structure} --- an abstract hierarchical structure that determines or constrains semantic composition --- and \emph{coding properties} --- expressive devices such as word order, function words and morphological inflection that are used to partially encode the syntactic structure.
To illustrate this point, let us consider two equivalent sentences in Finnish and English:

\enumsentence{\shortex{4}
{~koira & ~jahtasi & ~kissan & ~huoneesta}
{~dog-NOM & ~chase-PRS & ~cat-ACC & ~room-ELA}
{~`a/the dog chased a/the cat from a/the room'}
\label{fi}}

\enumsentence{~the dog chased the cat from the room \label{en}}

\noindent
Most linguists would agree that (\ref{fi}) and (\ref{en}) not only mean (roughly) the same thing but also have a similar syntactic structure, where the main verb (\emph{jahtasi}, \emph{chased}) takes a subject (\emph{koira}, \emph{the dog}), a direct object (\emph{kissan}, \emph{the cat}) and a locative modifier (\emph{huoneesta}, \emph{from the room}). However, the encoding of this syntactic structure is quite different in the two languages. In English, the subject and object are primarily identified through their position relative to the verb, while the locative modifier is introduced by a preposition (\emph{from}). In Finnish, the role of all three dependents of the verb is indicated by morphological case inflection, and constituent order is not significant.\footnote{In principle, the words of the Finnish sentence can be rearranged in any order without changing the syntactic roles, but some orders may be less natural and/or carry special pragmatic implications.} Note also that the overt coding properties (word order, function words, morphological inflection) do not (always) uniquely determine the syntactic structure. For example, in the English example, the phrase \emph{from the room} could also function as a modifier of the noun phrase \emph{the cat}, although this is a less likely interpretation in most contexts.

While coding properties are concrete aspects of the sentence, the syntactic structure is essentially an abstract concept that is not directly observable. Nevertheless, linguists have over the years accumulated compelling evidence for the existence of a hierarchical structure over and above the sequential order of words. The most obvious type of evidence is perhaps the occurrence of structural ambiguity, where a single sequence of words can be assigned multiple compositional interpretations, exemplified in the following classic examples:

\enumsentence{~she saw the man with the telescope \label{pp}}
\enumsentence{~old men and women \label{coord}}
\enumsentence{~flying planes can be dangerous \label{const}}

\noindent Other types of evidence come from substitution and permutation tests 
\citep[see, e.g., ][]{matthews81}. However, while the existence of a hierarchical structure is hardly contested today, the linguistic theories developed to account for this structure vary in their theoretical assumptions as well as in their mathematical representations of syntactic structure. A broad distinction can be made here between theories based on phrase structure (constituency) \citep{bloomfield33,chomsky57} and theories based on dependency structure \citep{tesniere59,melcuk88}, but there are also other approaches and considerable variation within each family of theories. To some degree, it is possible to convert syntactic representations from one theoretical framework to another, but the conversion is usually heuristic and lossy and, therefore, the different representations are not commensurable, strictly speaking. 

The existence of a wide range of syntactic theories arises from contested views on how a diverse range communicative principles, including the use of different coding properties, can come to exist across languages. For example, the Chomskyan 
tradition posits that an innate human grammar --- a set of rules and processes that govern human cognition --- is privy to a series of language-specific transformations that result in such idiosyncrasies \citep{chomsky65aspects,chomsky81,chomsky95}. Other accounts argue that syntax itself is shaped by functional or cognitive constraints \citep{zipf1949human,givon1995functionalism,hawkins2004efficiency,jaeger2011language,gibson2019efficiency}, such as managing memory load by preferring dependencies of shorter length \citep{gibson1998linguistic,gibson2000dependency} --- a process which can also influence coding properties like word order \citep{hahn2020universals,futrell2020dependency}. Cultural differences across languages are likewise theorized to play a large role \citep{tomasello2009cultural,evans2009myth}, with complex morphosyntactic processes like polysynthesis being largely observable in small, non-industrial communities with dense social-network structures \citep{trudgill2017anthropological}. Directly or not, such debates revolve around the controversial \textit{poverty of the stimulus} argument \citep{lasnik2017argument}--- linguistics' own spin on psychology's nature vs. nurture debate --- where the human capacity to acquire and generalize across structures is perceived as either predominantly learned or predominantly innate. 

Neural networks --- especially large scale language models --- have recently assumed an interesting place in this discussion.  Primarily, syntactic theory has offered a useful toolkit for more fine-grained evaluation of language models, which have shown an ability to generate coherent, grammatical output, resembling that of humans. To this end, researchers have employed well-studied coding properties like subject-verb agreement \citep{linzen16} or phenomena like filler-gap dependencies \citep{wilcox-etal-2018-rnn} to articulate exactly on which grounds a models' output might be judged as grammatical or not. Such studies have served as a welcome complement to the ubiqutious, yet opaque perplexity metric --- a measure of how predictable sentences or documents are, given a model's parameterization. In a sense, however, they can likewise be perceived as a means of \textit{sanity-checking} models' behavior \citep{baroni2021proper}, with paper titles often framed interrogatively: \textit{do neural language models learn} \underline{\hspace{0.75cm}}? Nonetheless, answering such questions is useful, 
and a concrete understanding of 
the ability of neural networks to generalize with respect to natural language --- as well as the algorithmic processes that underlie this capacity --- could, in the least, provide interesting perspectives on the age-old debates mentioned above \citep{linzen2021syntactic}.

\section{Review: The Quest for Syntax}
\label{sec:review}

In this section, we review work belonging to what we perceive as the three dominant paradigms for attesting language models' knowledge of syntax --- targeted syntactic evaluation, probing, and (downstream) NLU evaluation. Though comprehensive surveys of such studies can be found, for example, in \citet{linzen2021syntactic} or \citet{manning2020emergent}, our aim here is to relate them to the concepts and distinctions discussed in the previous section. Readers interested in a more detailed description and analysis are referred to the aforementioned work. 

\subsection{Targeted Syntactic Evaluation}\label{tse}
\noindent
Targeted syntactic evaluation\footnote{This term was coined, to the authors' best knowledge, by \citet{marvin2018targeted}.} (TSE) is arguably the most popular framework for assessing neural networks' ability to make syntactic --- therefore hierarchical --- generalizations. At its core, TSE is a black-box testing approach concerned with measuring model output (typically probabilities) with respect to a curated set of stimuli. Such stimuli are typically based on minimal pairs motivated by phenomena in the syntax literature. For example, consider the (by now classic) example in sentences \ref{tempa} and \ref{tempb}.

\eenumsentence{
    \item{~The keys to the cabinet are on the table. \label{tempa}}
    \item{~*The keys to the cabinet is on the table. \label{tempb}}
    \item{~*The key to the cabinets are on the table. \label{tempc}}
    \item{~The key to the cabinets is on the table. \label{tempd}}}\label{tempfull}



\noindent
The literature dictates that a competent English speaker would rely on a structural analysis of \emph{the keys to the cabinet} to infer number agreement between the plural subject (\emph{keys}) and the copula verb (\emph{are}). On the other hand, a purely 
sequential processing of the sentence would arrive at the opposite conclusion in \ref{tempb}: \emph{is} agrees with the adjacent singular noun (\emph{cabinet}). To ascertain whether or not a language model $M$ follows the former logic, 
one could, for example, compare the probabilities assigned to the target verb \textit{be} in \ref{tempa}-\ref{tempb}, given the context $C = $ \emph{the keys to the cabinet}, and examine whether $P_{M}(\mathrm{are}|C) > P_{M}(\mathrm{is}|C)$. This can also be extended to full paradigms, where, in the case of \ref{tempfull}, $M$ has to assign higher probabilities to both (\ref{tempa}) and (\ref{tempd}) with respect to (\ref{tempb}) and (\ref{tempc}). TSE (per this formulation) can thus be seen as based on an accuracy metric, which, if returning a high value over $n$ stimuli, implies that $M$ is able to generalize with respect to the relevant syntactic phenomenon. 
It should be noted that probability assigned to the word form $x$, per various theoretical justifications, is sometimes replaced with surprisal, e.g., $S = -\mathrm{log}_{2}P_{M}(x|C)$, as in \citet{wilcox-etal-2018-rnn} and \citet{futrell2019neural}. Furthermore, in situations where the locus of ungrammaticality does not lie on a single word (as in English subject-verb agreement), but is dependent on the interaction of several words (e.g., as in negative polarity items), it is common to compare the probabilities or perplexities of entire sentences \citep{marvin2018targeted,jumelet-hupkes-2018-language}.

The TSE framework also allows for flexibility in integrating more complex sets of stimuli, as in the study on syntactic state by \citet{futrell2019neural}:

\eenumsentence{
    \item{~As the doctor studied the textbook, the nurse walked into the office. \label{futa}}
    \item{~*As the doctor studied the textbook. \label{futb}}
    \item{~?The doctor studied the textbook, the nurse walked into the office. \label{futc}}
    \item{~The doctor studied the textbook. \label{futd}}}\label{futfull}
\noindent
With respect to (\ref{futfull}), \citet{futrell2019neural} formulate a set of hypotheses, whereby they posit (1) that the surprisal at the matrix clause after the comma (... \emph{the nurse walked into the office}.) should be lower for (\ref{futa}) than for (\ref{futc}) (the network knows it is in a subordinate clause per the subordinator \emph{as}), and (2) that the surprisal at the matrix clause should be higher for \ref{futb} than \ref{futd} (the network expects a matrix clause per the subordinator). Though the aforementioned accuracy approach could likewise be appropriated here as a summary statistic, researchers also often employ 
significance testing in order to accept or reject their hypotheses. For example, \citet{futrell2019neural} apply a linear mixed-effect model on their models' stimulus-level predictions in order to accept hypothesis (1) on behalf of all models, but reject hypothesis (2) for all but two. This formulation --- in line with common paradigms in psycholinguistics --- leads them to conclude that, while all models are partially capable of tracking syntactic state across subordinate and main clauses, certain training conditions are required (large data or explicit structural objectives) in order to fully capture the structural expectations induced by subordinators. A similar methodology is employed in \citet{wilcox-etal-2018-rnn} for investigating filler-gap dependencies. 

The popularity of the TSE framework has precipitated the creation of challenge suites, which offer holistic measures of models' performance across a variety of linguistic phenomena.  \citet{marvin2018targeted} were among the first to introduce such datasets, employing a context-free grammar to procedurally generate minimal pair sentences --- such as \ref{tempa} and \ref{tempb} --- for a variety of phenomena: agreement (of various kinds), reflexive anaphora, and negative polarity items. \citet{warstadt2020blimp} later presented a similar, automatically generated dataset of minimal pairs (BLiMP), albeit with wider coverage: 1000 sentences per 67 paradigms belonging to 12 different phenomena. The authors used BLiMP to study various popular language model architectures (LSTM, Transformer), whereby they associated average accuracy across phenomena with models' linguistic knowledge. A similar suite was contemporaneously introduced by \citet{hu2020systematic}, albeit in employ of $2\times2$ templates like \ref{tempfull} for hand-curated stimuli culled from syntax textbooks. Like \citet{warstadt2020blimp}, \citet{hu2020systematic} used their suite\footnote{This suite was later named \textsc{SyntaxGym} in \cite{gauthier2020syntaxgym}.} to study language model architectures, most notably relating language models' syntactic generalization (SG) score --- measured in aggregate across phenomena --- to their test set perplexity.

\subsection{Probing}
\noindent
Probing\footnote{Also known as diagnostic classification \citep [see, e.g.,][]{hupkes18}.} is another popular paradigm for attesting NLP models' acquisition of syntax. The key distinction between TSE and probing is that, while the former is concerned with model behavior, the latter focuses explicitly on model \textit{representation}. In this context, behavior is likened to the probabilities assigned to certain outputs (extracted, typically, from the output layer of a language model), while representation refers to the intermediate hidden state vectors computed by the model. Mainly, probing is motivated as being necessary due to deep learning's end-to-end nature: features are learned with respect to a given task, not engineered like in traditional systems. Due to this fact, neural models' representations are wholly uninterpretable to the human interlocutor and thus require intervention in order to understand what they portray.

Formally speaking, probing is concerned with representations $\mathbf{h}$ extracted from a model $M$ for a given input $x$: $\mathbf{h} = M(x)$. A representation $\mathbf{h} \in \mathbb{R}^{1 \times d}$ is typically a fixed-length dense vector corresponding to input word $x$ (e.g., \emph{keys} in \ref{tempa}), where $d$ is the hidden-state dimensionality of $M$. A probe $f$ for a given linguistic property $A$ is a classifier fit on $\mathbf{h}$ to produce output $y \in Y$, where $Y$ is a finite label set: $y = f_{A}(\mathbf{h})$. For properties that can be decoded from single words, such as part-of-speech (POS) tags, a trained probe $f_{\mathrm{POS}}$ must be able to assign the correct label to $\mathbf{h}$ with respect to the ground truth, e.g., $\hat{y} = \mathrm{NOUN}$ for $M(\mathrm{keys})$. For properties concerning two or more words, such as dependencies or phrases, a concatenation of hidden states corresponding to discontiguous tokens $x_{i}, x_{j}$ or a contiguous span of tokens $x_{i},...,x_{j}$ is applied. In this latter formulation, deemed edge-probing by \citet{tenney2019you}, one might expect a probe $f_{\mathrm{DEP}}$ to decode $\hat{y} = \mathrm{NSUBJ}$ for $M(\mathrm{keys},\mathrm{are})$ and $f_{\mathrm{CON}}$ to decode $\hat{y} = \mathrm{PP}$ for $M(\mathrm{on},\mathrm{the},\mathrm{table})$. Though probing models vary widely in terms of architecture, parameters, optimization, etc., the vast majority of them assume a training set $D_{A}$ representative of property $A$ on which $f$'s parameters $\Theta$ can be fit, like a treebank. Such probes are then evaluated in standard supervised learning fashion via accuracy on a held out test set.
 If such accuracy is high, it can then be said that $A$ is decodable from $\mathbf{h}$, i.e. that $M$ learns it. This framework was notably employed by \citet{liu-etal-2019-linguistic}
and \citet{tenney2019you}, who concurrently demonstrated that representations extracted from popular contextual embedding models (ELMo, BERT, GPT) yielded exceedingly good performance on suites of linguistic tasks. Also noteworthy is \citet{tenney-etal-2019-bert}'s study, which showed that BERT's representations appear to evolve in capturing properties with increasing levels of complexity, from lexical features to syntax and semantics. 

While the aforementioned word-level approach is by far the most popular probing setup, other methods for decoding the syntactic structure of entire sentences have been proposed. One model that is of particular interest is that of \citet{hewitt19}, who attempt to decode dependency structure from models' vector spaces. To this end, they propose to learn transformations over model representations, such that (1) the squared $l_2$ distance between any vectors $\mathbf{h}_{i},\mathbf{h}_{j}$ reflects the distance between their corresponding words $x_i, x_j$ in a parse tree, and (2) that the $l_2$ norm of any vector $\mathbf{h}_{i}$ reflects the depth of its corresponding word $x_i$ in a parse tree. They find that this method is particularly effective for decoding Stanford Dependencies trees \citep{de-marneffe-etal-2006-generating} from ELMo and BERT representations, with respect to several lexical-only baselines. Beyond \citet{hewitt19}'s method, which can be imagined as doing parsing by proxy, other work has directly employed (underparameterized) dependency parsers as probes. For example, see \citet{hewitt-liang-2019-designing}, who employ a graph-based bilinear probe; \citet{maudslay2020tale}, who investigate the relation between probing and parsing; and \citet{pimentel-etal-2020-pareto}, who advocate for adding full dependency parsing to the probing task suite. 

At this stage, probing can be considered a field of inquiry in its own right, with researchers presenting new models, metrics, and criticisms for every conference cycle. Naturally, the use of intermediary models trained on top of extracted representations warrants caution from the interlocutor. Some concerns expressed in the literature include, but are not limited to: the use of smaller, linear models vs. larger, nonlinear ones; appropriate baselines and evaluation metrics; properties being learned by the probe vs. occurring in representations; properties being employed by the model in the original task vs. simply being decodable, etc. Though a full consideration of these methodological concerns is outside the scope of this article, we refer the interested reader to \citet{belinkov2021probing} --- a comprehensive review of the paradigm, open issues, and alternative approaches like attention analysis.

\subsection{NLU Evaluation}
\noindent
Outside of TSE and probing, another technique that has recently attracted much attention is the evaluation of models (imbued with or deprived of syntactic knowledge) on downstream tasks. The logic inherent to this line of inquiry is as follows: if a model has come to rely on human-like knowledge of language to solve complex NLP tasks, then it should (1) perform \textit{poorly} on such tasks when the surface form of an utterance has been corrupted beyond (human) comprehension, and (2) perform \textit{identically} when imbued with the exact abstract structure theorized by linguists as governing the surface form. Such tasks are typically taken from the GLUE benchmark --- a suite of natural language understanding (NLU) datasets ``designed to favor and encourage models that share general linguistic knowledge across tasks'' \citep{wang2018glue}. GLUE has served as the principal point of comparison for pretraining architectures, where, as of writing, 15 models have surpassed the published human performance on the same tasks. 

In terms of input corruption, many studies have investigated the effect of word order on NLU task performance. Indeed, word order is the primary means of encoding syntactic argument structure in English, and such work often hypothesizes that sensitivity to this particular property should result in lower NLU scores. \citet{gupta2021bert} demonstrate that this is not the case for BERT when fine-tuning on various GLUE tasks: sequences corrupted at test-time by means of shuffling, sorting, duplicating, and dropping tokens still retain 70-90\% performance of the non-perturbed input. Moreover, models appear to be as confident in assigning labels to perturbed inputs as they are to naturalistic ones. These results are corroborated by \citet{pham2020out}, who show that models predominantly seek salient words in sequences, with numerous attention heads specializing themselves for this exact purpose. 
\citet{sinha2020unnatural} report similar findings for various NLI datasets (in English and Chinese) across a variety of model architectures. They show that models are insensitive to word reorderings, some of which can actually result in improved task performance. Perhaps most strinkingly, \citet{sinha21} show that \textit{pre-training} full-scale RoBERTa models on perturbed sentences (across n-grams of varying lengths) and fine-tuning them on unaltered GLUE tasks leads to negligible performance loss. They also report that a popular probe for dependency structure, that of \citet{pimentel-etal-2020-pareto}, is able to decode trees from the perturbed representations --- even a unigram baseline with resampled words --- with considerable accuracy.

As a conceptual counterpoint to the permutation-based line of research, several studies have posed the opposite question: does explicitly injecting syntactic structure into models' representations or training objectives lead to better downstream performance? The observations in such studies are similar to the aforementioned work, albeit slightly more subtle: models that factor syntax into their decisions generally do not benefit in performance via its injection, which is taken to imply that such structure is redundant to the model, or not needed at all. Most notably, \citet{glavavs2021supervised} fine-tune BERT and RoBERTa \citep{liu2019roberta} as dependency parsers, before fine-tuning the same models again on NLI, paraphrase detection, and commonsense reasoning tasks. They show that, while intermediate parsing training (IPT) can produce near state-of-the-art parsers, repurposing these parameters for NLU tasks leads to negligible improvement. A similar trend is shown in \citet{kuncoro2020syntactic}, who train a BERT model distilled from an RNNG teacher \citep{dyer-etal-2016-recurrent}. They, too, find that, while their syntactically-aware model achieves top marks on a suite of parsing and otherwise syntactic tasks, the benefits for fine-tuning on GLUE are scant, if any. \citet{swayamdipta2019shallow} corroborate these findings for 
ELMo models conditioned on chunked input derived from phrase structure trees.

\section{Discussion: A Call For Clarity and Caution}
\label{sec:discussion}

After our general discussion of syntax, as well as our review of work exploring its role in contemporary language models, we are now in a position to make a few basic distinctions. In this section, we attempt to situate the findings of the aforementioned studies along several dimensions that we deem important towards the advancement of our research agenda.

\subsection{Coding Properties are not Syntax}
\noindent
First, we would like to highlight the need to be clear about whether a study is concerned with abstract syntactic structure, overt coding properties, or with some relation between the two. A typical fallacy that may arise from not observing this distinction is to conflate a particular coding property with the abstract syntactic structure that it partially encodes. Naturally, if we fall victim to this fallacy when interpreting certain findings, we risk drawing conclusions based on insufficient or irrelevant evidence. This applies to situations where we may be tempted to employ coding properties as proxies of syntactic structure --- either for attesting models' sensitivity to the latter or refuting it. 

For example, it is important to acknowledge that studying agreement via TSE gives us a glimpse into how language models capture the syntactic relationship between selected words, such as verbs and their subjects. Per this view, high performance --- even in the presence of various types of attractors --- does not necessarily ential that a model has learned the grammar of a language. Rather, it has simply shown itself to be particularly sensitive to a single coding property, grammatical relation, or dependency type. Notably, English agreement is limited to expressing the number or person of the subject on the finite main verb (when in the present tense). This amounts to being, in the vast majority of cases, a binary distinction between correct and incorrect inflections, which bears a strong random choice baseline of 50\% in the case of TSE. Thus, when one considers types of agreement manifested in other languages --- such as number, gender, and case agreement between nouns and their modifying adjectives (e.g., German, Russian), or polypersonal agreement between a verb and multiple arguments (e.g.,  Basque, Georgian) --- it becomes difficult to judge agreement as the primary mechanism by which syntax is encoded in English. Indeed, studies have shown that models tend to struggle with more expressive agreement mechanisms in morphologically rich languages \citep{ravfogel-etal-2018-lstm}. Such insights call not only for a typologically driven research agenda, but also for nuance in interpreting positive findings for singular properties in selected languages.\\

We must also note that the above logic can apply in reverse: a model's lack of sensitivity to a single coding property, for example, word order \citep{dryer1992greenbergian}, does not imply that the model has failed to acquire syntax as a byproduct of its training objective. 
Even in a language like English, where word order is very salient, it is not the only coding property that signals syntactic structure. Consider \emph{chases the cats the dog} as a permutation of \emph{the dog chases the cats}: it is not unreasonable for an English speaker to decode the argument structure of this permutation using subject-verb agreement alone. Indeed, recent research in psycholinguistics has intimated that humans are relatively robust to permutations of linguistic form \citep{traxler2014trends}. In the context of word order, \citet{mollica2020composition} show that humans are able to process permuted sentences similarly to naturalistic ones, albeit when local structure (measured via pointwise mutual information) is preserved. Recently, this has been corroborated for models fine-tuned on GLUE as well, with performance therein strongly correlated with the extent of local structure corruption \citep{clouatre2021demystifying}. With this in mind, one can see that order perturbation studies do not provide enough evidence to conclude that models (or humans) are insensitive to syntax. Instead, when conducting such studies, we must recall that word order (or agreement, for that matter) is simply a single coding property in a mosaic of such properties, all of which are privy to underlying processes that drive composition and comprehension. 
\\

\subsection{Syntactic Representations are not Linguistic Data}

As a second point, if a study is concerned with syntactic structure, we need to clarify whether it assumes a specific type of syntactic representation, since the choice of representation may affect the results. Other things being equal, we may therefore prefer methods that do not presuppose specific syntactic representations, since conclusions will otherwise be valid only on the assumption that the chosen representation correctly captures syntactic structure. This consideration is even more important when we make use of automatically parsed data --- as opposed to manually annotated sentences from treebanks --- where otherwise sound syntactic representations may give misleading results due to parsing errors. At the same time, it is important to note that avoiding syntactic representations altogether may be limiting in another way, as it may restrict our methodological repertoire. Thus, as long as we maintain a critical attitude towards representation-dependent methods, they may still provide us with valuable results that cannot be obtained with other methods.

To illustrate the importance of representations in the context of probing, we can start by asking: does high UAS on a particular treebank imply that those trees are indeed \textit{the} structures encoded by a given model? Or can alternative, linguistically plausible structures be decoded with comparable accuracy? \citet{kulmizev20} explore this question when probing various models for UD, a dependency formalism which prioritizes content-word heads \citep{demarneffe21}, and Surface-Syntactic UD, which assumes a traditional function-word head style analysis \citep{gerdes2018sud}. They find that, while the difference in decoding UAS between the two formalisms is minimal for some treebanks, other treebanks exhibit strong preferences for either UD or SUD. They contribute such preferences to a complex interplay between the formalisms' inherent graph properties (e.g., average tree height), the probe employed for decoding (\citet{hewitt19}'s, in their case), annotation factors like tokenization, and morphology. Though preliminary, \citet{kulmizev20}'s study is a cautionary tale in tree-based probing, where choice of representation directly affects what conclusions one may draw about models.

We can ask similar questions when attempting to imbue models with syntactic structure. For example, is the injection of UD trees into a model's architecture enough to draw conclusions about the role of syntax in downstream performance? Or do alternative, linguistically plausible representations exist that models might yet benefit from? Beyond this, what privileges one particular injection method, say intermediate parsing training \citep{glavavs2021supervised}, over another, such as knowledge distillation from an RNNG teacher \citep{kuncoro2020syntactic}? A template for exploring such considerations can be found in \citet{wu2021infusing}, who report that infusing BERT with \textit{semantic} dependencies can provide modest gains on GLUE. In that study, they compare the DM representation focused explicitly on predicate-argument structure  \citep{ivanova2012did} with the more syntactically oriented UD, finding that the former leads to slightly better performance.\footnote{Interestingly, both syntactic and semantic models seem to outperform the fine-tuned RoBERTa baseline.} Furthermore, they compare their chosen infusion method --- semantic graph embeddings learned via a relational graph convolution encoder \citep{schlichtkrull2018modeling} --- with other means of injecting structure into representations, where their method performs best in most cases. 




\subsection{Theory, Model, and Task}

As a further means of promoting clarity, we consider it vital, for any study investigating the syntactic capabilities of language models, to motivate the evaluation with a direct aspect of syntax in mind. Indeed, TSE studies on coding properties and probing studies on tree representations satisfy this desideratum by design. Studies that investigate models' knowledge of syntax through the prism of downstream tasks, however, do not. 

For example, if we observe models achieving high performance on NLU tasks, we may be tempted --- by virtue of the NLU branding --- to assume that they are, in some fashion, similar to humans in their decision-making. To ascertain the extent to which this is true, we must first attempt to articulate exactly how humans behave when choosing a particular label for a sample over another. Indeed, this has, to an extent, been achieved for some textual entailment datasets, where annotators were asked to justify their labeling decisions with free-text \textit{rationales} \citep{camburu2018snli,rajani-etal-2019-explain}. Such datasets, however, introduce a third variable into the mix (input, output, and explanation), which complicates architecture design and further divorces the original model (one that may be evaluated, presumably, for its syntactic capabilities) from the task on which it is now being fine-tuned and offering explanations for. Furthermore, and perhaps more importantly, the existence of a third variable complicates the interpretation of the self-rationalizing model itself, as noted by \citet{wiegreffe2020measuring}: for example, how might we ensure that the model's explanation is indeed faithful to the label it produced? To its input?

Given that we typically lack an explicit characterization of how humans make labeling decisions at the sample level, we are then left to interpret the behavior of the model itself. For example, if working with Transformer models, we may be tempted to examine their self-attention weights after fine-tuning, as in \citet{clark2019does} or \citet{htut2019attention}. However, the prospect of treating attention as explanation has received its fair share of criticism in recent years, due to (1) a lack of input token identifiability in deep Transformer models \citep{brunner2019identifiability,abnar-zuidema-2020-quantifying} and (2) an unclear correspondence between a model's attention patterns and its outputs \citep{jain-wallace-2019-attention,wiegreffe-pinter-2019-attention,serrano-smith-2019-attention}. Due to these drawbacks, we may then attempt to employ alternative methods of explanation, such as gradient-based saliency \citep{simonyan2013deep,sundararajan2017axiomatic}, Shapley value-based feature attributions \citep{lundberg2017unified}, or model agnostic approaches \citep{ribeiro2016should}, inter alia. Here, we are met with a space of methods, whose utility in NLP tasks has been surveyed with mixed results \citep{poerner-etal-2018-evaluating,arras-etal-2019-evaluating,atanasova-etal-2020-diagnostic}. Hence, in order to draw insights about a model's syntactic capabilities vis-a-vis its performance on a downstream task, we must not only make a principled choice of explanation model, but also relate that model's explanations --- which are computed with respect to predictions downstream --- to the particular aspect of syntax we are interested in. Again, questions central to \textit{faithfulness} \citep{jacovi2021aligning} become vital: how can we guarantee that an explanation of our model's behavior represents what it actually did? Further yet, how can we use explanations pertaining to one type of behavior to shed light on another, unobserved one?

Certainly, numerous caveats are endemic to the process of interpreting model behavior \citep{lipton2018mythos}. This may lead us to put our full trust in the \textit{task}, which we can employ as a prism through which to opine on the model. This necessitates that the task is indeed well-motivated and designed, and difficult to exploit via heuristics. If we believe this to be true, we can hypothesize that, by performing well on such tasks, our models possess whatever latent ability humans do in solving them --- see, e.g., \citet{sinha2020unnatural}: ``models should have to know the syntax first, \ldots if performing any particular NLU task that genuinely requires a humanlike understanding of meaning''. Unfortunately, in the context of NLI (which \citet{sinha2020unnatural} study) this is a highly dubious claim: the crowd-funded nature of such datasets makes them prone to annotation artefacts (e.g., subsequence overlap between premise/hypothesis, lexical choice across inference classes, sentence length, etc.), which models often exploit as heuristics, thus leading to highly inflated performance metrics \citep{gururangan-etal-2018-annotation,poliak-etal-2018-hypothesis,mccoy-etal-2019-right}. 

Ultimately, the extent of trust we place in the model (performing as hypothesized) over the task (being correctly expressed) may influence not only our hypotheses, but also the conclusions we draw from our findings. For example, consider \citet{pham2020out} as a counterpoint to \citet{sinha2020unnatural}. Though the observations regarding BERT-based models' insensitivity to word order are largely similar, the former are more critical of the task (``GLUE does not necessarily require syntactic information or complex reasoning''), and the latter of the model (``current models do not yet `know syntax' in the fully systematic and human-like way we would like them to''). Naturally, for this and the aforementioned reasons, disentangling the complex web of theory, model, and task remains a gargantuan undertaking. However, if we formulate our hypotheses using the exact behavior we would like to investigate and evaluate our models along those same lines, then the conclusions we draw from our findings should be much clearer to interpret.

\subsection{What are the Research Questions?}

In addition to specifying which aspect of syntax a study is concerned with, we also need to be clear about what our research questions are. For example, given a model $M$, a task $T$, some training data $D$ and some aspect of syntax $A$, we may ask the following three questions:
\begin{enumerate}
    \item To what degree \emph{does} $M$ learn $A$ when trained on $D$ to perform $T$?
    \item To what degree \emph{can} $M$ learn $A$ when trained on $D$ to perform $T$?
    \item To what degree does $M$ \emph{need} to learn $A$ when trained on $D$ to perform $T$?
\end{enumerate}
Questions of type 1 are the most straightforward to investigate as long as we have a valid and reliable method for measuring the degree to which $M$ learns $A$ in the context of $D$ and $T$. This is quite a big assumption in itself, and one that we will return to shortly, but we will focus first on the logic for answering different research questions. Questions of type 2 are 
modal in nature and therefore hard to investigate empirically, except indirectly by investigating questions of type 1. For example, in the pioneering study by \citet{linzen2016assessing}, discussed in Section~\ref{sec:review}, the authors were primarily interested in whether an LSTM ($M$) can learn ``syntax-sensitive dependencies'' ($A$) --- a question of type 2. To investigate this, they examined the actual learning behavior of the model in two specific settings (questions of type 1): (a) when trained on unlabeled text ($D_U$) for the task of language modeling ($T_{LM}$), and (b) when trained on labeled sentences ($D_L$) for a specific agreement decision task ($T_{A}$). The results were largely negative in the first case and positive in the second. From the positive result, they could conclude that the model \emph{can} learn the relevant dependencies when trained on $D_L$ for $T_A$; from the negative results, they could however only conclude that there was no evidence that the model was capable of learning the relevant aspect of syntax when trained on $D_U$ for $T_{LM}$. This illustrates the fundamental asymmetry between positive and negative results when it comes to generalizations about possibility. A single positive result --- if interpreted correctly --- is sufficient to establish that something is possible, while any number of negative results are in principle inconclusive.\footnote{This is the mirror image of the case of necessity, underlying the famous quotation attributed (probably incorrectly) to Einstein: ``No amount of experimentation can ever prove me right; a single experiment can prove me wrong.''} Indeed, as discussed in Section~\ref{sec:review}, the later study by  \citet{gulordava2018colorless} managed to obtain positive results also in a setting similar to the first scenario of \citet{linzen2016assessing}, from which they concluded that LSTMs are capable of learning at least one aspect of syntactic structure without explicit supervision.  

Questions of type 3 are more complex still, because they involve causality as well as modality. More precisely, they combine the question of whether learning $A$ results in better performance of $M$ on $T$ (causality) with the question of whether learning $A$ is necessary to achieve better peformance (modality). A typical example is the study of \citet{glavas21}, discussed in Section~\ref{sec:review}, where the authors study the effect of intermediate parser training of a pre-trained language model later fine-tuned for various language understanding tasks. The underlying research question is whether knowledge of syntax is needed for language understanding --- a question of type 3 --- and the lack of improvement may suggest a negative answer, but this conclusion is only warranted if it can also be shown (a) that the model has actually learned (some aspects of) syntax and (b) that this knowledge causally affects the model's behavior on the downstream task (and still fails to improve performance). Note, however, that a positive improvement would not be more conclusive in this case, because it would only show that improvement is \emph{possible}, not that it is \emph{necessary}. This illustrates the complexity involved when relating experimental results to research questions and again points to the need for careful meta-analysis.

\subsection{Aggregate Metrics may be Misleading -- but are Necessary}
Let us finally turn to the question of how we can measure the degree to which a model $M$ learns some aspect of syntax $A$ when trained on data set $D$ to perform task $T$ --- a question that is crucial to all studies in this area, regardless of what the more general research questions are. As we have seen in Section~\ref{sec:review}, the answer usually involves measuring performance on an appropriate task $T'$, although the exact solution depends on the type of study. In TSE studies, $T'$ is typically the task of discriminating positive from negative instances of some grammatical pattern, for example by assigning higher probability to the positive instance in a minimal pair. In probing, $T'$ is a supervised classification task assumed to reflect syntactic knowledge. And in NLU evaluation, $T'$ is simply the downstream language understanding task and thus normally coincides with the main task $T$. Each of these paradigms comes with its own methodological pitfalls, which have been extensively discussed in particular in the case of probing, but we will focus here on the complexities that are common to all of them.

First of all, we note that performance on $T'$ is almost always measured by averaging over individual test instances. In the simplest case, this may just be the arithmetic mean of a 0-1 loss metric, such as the accuracy reported for a probing classifier predicting part-of-speech tags. In other cases, it may be a more or less sophisticated macro-average, like an average over different grammatical patterns in a TSE study. In all cases, however, such aggregate measures need to be interpreted carefully. 
First of all, how do we know whether a given metric value indicates presence or absence of syntactic knowledge? Does a value of 0.5 mean that the glass is half full or half empty? This highlights the importance of relevant and informative baselines, a point that has been made in the literature before but that has perhaps not been fully appreciated.

Second, it is in the nature of aggregate metrics that they can easily be misleading by hiding important variation, especially if the distribution of different types of phenomena is heavily skewed. For example, in the related field of syntactic parser evaluation, \citet{rimell09} have shown that parsers with very respectable performance according to standard aggregate metrics like EVALB can have close to zero accuracy on certain types of unbounded dependency constructions. Moreover, aggregation may hide important variation in a number of different ways. If we use naturally occurring text in our test sets, certain words and constructions will inevitably be much more frequent than others and therefore dominate the aggregate scores in the same way as for syntactic parser evaluation. As a result of this, \citet{newman21} argue that standard metrics used in TSE overestimate the systematicity of language model behavior. If in addition we aggregate over different syntactic phenomena, we may hide the fact that different phenomena are captured to different degrees. And if we aggregate over multiple languages --- or only report results for a single language --- we may neglect important language-specific properties and risk over-generalization.

Lastly, we must consider the role aggregation plays in the interpretation of models' performance on benchmarks like BLiMP, SyntaxGym, or GLUE. At its core, such an enterprise entails that all aspects of syntax or language understanding --- at least those of particular salience --- have been successfully enumerated. Given the abstract nature of these notions, and the extent of debate regarding them, it is naturally doubtful that such an enumeration could ever be attained. Relaxing this somewhat, in assuming a salient set of aspects has indeed been collected, one must likewise assume --- before aggregating --- that a principled weighting of such aspects exists. This is especially relevant when dealing with a space of tasks or phenomena where fine-grained categorizations are likewise included --- for instance, the 6 subject-verb agreement settings attested in BLiMP. In such cases, one must not only choose between micro or macro averaging across phenomena and their fine-grained attestations, but also articulate whether or not all phenomena lie on an equal playing field --- in other words, that they are all equally (1) difficult to attest and (2) salient for evaluation. Certainly, in the vast majority of cases we assume a uniform weighting of classes when aggregating, since introducing hand-selected weights may introduce bias that we would otherwise prefer to avoid. However, we must not fail to acknowledge that benchmarks, in themselves, are influenced by designers' theories on what component parts adequately represent abstract notions like syntax or language understanding. 

There is unfortunately no simple remedy to the complexities discussed in this section. In particular, giving up aggregate metrics is definitely not an option, since they are necessary for statistical significance testing and generalization. However, to make further progress in this area, we need to be aware of their inherent limitations, be open to consider alternative metrics, and to make sure to complement them with more specific forms of analysis whenever possible. 

\section{Conclusion}
\label{sec:conclusion}

The rapid progress in NLP thanks to deeper and larger neural network models trained on very large data sets with little or no linguistic supervision raises a number of questions concerning the status of traditional linguistic notions and theories in this landscape. Is there still a role for traditional techniques like supervised syntactic parsing? If not, is this because neural language models learn the relevant generalizations about linguistic structure without explicit supervision, or because language understanding does not really depend on such generalizations in the way traditionally assumed. If the latter, does this hold only for language understanding by machines, or does it also have implications for human language understanding?

These are exciting questions and it is therefore not surprising that we have seen a considerable body of research in this area recently. They are also hard questions, and the methodology for tackling them is still under development, so it is also not surprising that results so far have been inconclusive and sometimes contradictory. As stated in the introduction, the goal of this article has not been to criticize previous efforts, but to contribute to our understanding of methods and results by articulating and discussing some of the inherent complexities in this research area. Without pretending to have any solutions, we want to conclude with some recommendations for future research, echoing the main points made throughout the paper. 

Our general recommendation can be summed up as a plea for clarity and caution. Clarity means being clear about what conception of syntax underlies our investigations, which aspects of syntax are being studied, and whether we make specific assumptions about syntactic representations. Clarity also implies that we explicitly discuss what research questions are being asked, and how they can be elucidated by the specific experiments we perform. Caution means being careful when interpreting aggregate results, always looking for alternative metrics and additional analysis, and making sure to consider evidence from multiple languages if we want to draw conclusions about natural language in general. Caution finally means resisting the temptation to draw strong conclusions from any single study, which is usually impossible given the complex interplay of research questions, methodology and data. It is only by piecing together all available evidence that we can hope to see the forest for the trees.

\bibliography{main}

\begin{thebibliography}{98}
\expandafter\ifx\csname natexlab\endcsname\relax\def\natexlab#1{#1}\fi

\bibitem[{Abnar and Zuidema(2020)}]{abnar-zuidema-2020-quantifying}
Abnar, Samira and Willem Zuidema. 2020.
\newblock Quantifying attention flow in transformers.
\newblock In \emph{Proceedings of the 58th Annual Meeting of the Association
  for Computational Linguistics}, pages 4190--4197, Association for
  Computational Linguistics, Online.

\bibitem[{Arras et~al.(2019)Arras, Osman, M{\"u}ller, and
  Samek}]{arras-etal-2019-evaluating}
Arras, Leila, Ahmed Osman, Klaus-Robert M{\"u}ller, and Wojciech Samek. 2019.
\newblock Evaluating recurrent neural network explanations.
\newblock In \emph{Proceedings of the 2019 ACL Workshop BlackboxNLP: Analyzing
  and Interpreting Neural Networks for NLP}, pages 113--126, Association for
  Computational Linguistics, Florence, Italy.

\bibitem[{Atanasova et~al.(2020)Atanasova, Simonsen, Lioma, and
  Augenstein}]{atanasova-etal-2020-diagnostic}
Atanasova, Pepa, Jakob~Grue Simonsen, Christina Lioma, and Isabelle Augenstein.
  2020.
\newblock A diagnostic study of explainability techniques for text
  classification.
\newblock In \emph{Proceedings of the 2020 Conference on Empirical Methods in
  Natural Language Processing (EMNLP)}, pages 3256--3274, Association for
  Computational Linguistics, Online.

\bibitem[{Baroni(2021)}]{baroni2021proper}
Baroni, Marco. 2021.
\newblock On the proper role of linguistically-oriented deep net analysis in
  linguistic theorizing.
\newblock \emph{arXiv preprint arXiv:2106.08694}.

\bibitem[{Belinkov(2021)}]{belinkov2021probing}
Belinkov, Yonatan. 2021.
\newblock Probing classifiers: Promises, shortcomings, and alternatives.
\newblock \emph{arXiv preprint arXiv:2102.12452}.

\bibitem[{Bloomfield(1933)}]{bloomfield33}
Bloomfield, Leonard. 1933.
\newblock \emph{Language}.

\bibitem[{Brunner et~al.(2019)Brunner, Liu, Pascual, Richter, Ciaramita, and
  Wattenhofer}]{brunner2019identifiability}
Brunner, Gino, Yang Liu, Damian Pascual, Oliver Richter, Massimiliano
  Ciaramita, and Roger Wattenhofer. 2019.
\newblock On identifiability in transformers.
\newblock \emph{arXiv preprint arXiv:1908.04211}.

\bibitem[{Camburu et~al.(2018)Camburu, Rockt{\"a}schel, Lukasiewicz, and
  Blunsom}]{camburu2018snli}
Camburu, Oana-Maria, Tim Rockt{\"a}schel, Thomas Lukasiewicz, and Phil Blunsom.
  2018.
\newblock e-snli: Natural language inference with natural language
  explanations.
\newblock \emph{arXiv preprint arXiv:1812.01193}.

\bibitem[{Chomsky(1957)}]{chomsky57}
Chomsky, Noam. 1957.
\newblock \emph{Syntactic Structures}.

\bibitem[{Chomsky(1965)}]{chomsky65aspects}
Chomsky, Noam. 1965.
\newblock \emph{Aspects of the Theory of Syntax}.

\bibitem[{Chomsky(1981)}]{chomsky81}
Chomsky, Noam. 1981.
\newblock \emph{Lectures on Government and Binding}.

\bibitem[{Chomsky(1995)}]{chomsky95}
Chomsky, Noam. 1995.
\newblock \emph{The Minimalist Program}.

\bibitem[{Clark et~al.(2019)Clark, Khandelwal, Levy, and
  Manning}]{clark2019does}
Clark, Kevin, Urvashi Khandelwal, Omer Levy, and Christopher~D Manning. 2019.
\newblock What does bert look at? an analysis of bert’s attention.
\newblock In \emph{Proceedings of the 2019 ACL Workshop BlackboxNLP: Analyzing
  and Interpreting Neural Networks for NLP}, pages 276--286.

\bibitem[{Clouatre et~al.(2021)Clouatre, Parthasarathi, Zouaq, and
  Chandar}]{clouatre2021demystifying}
Clouatre, Louis, Prasanna Parthasarathi, Amal Zouaq, and Sarath Chandar. 2021.
\newblock Demystifying neural language models' insensitivity to word-order.
\newblock \emph{arXiv preprint arXiv:2107.13955}.

\bibitem[{Devlin et~al.(2019)Devlin, Chang, Lee, and Toutanova}]{devlin19}
Devlin, Jacob, Ming-Wei Chang, Kenton Lee, and Kristina Toutanova. 2019.
\newblock {BERT:} pre-training of deep bidirectional transformers for language
  understanding.

\bibitem[{Dryer(1992)}]{dryer1992greenbergian}
Dryer, Matthew~S. 1992.
\newblock The greenbergian word order correlations.
\newblock \emph{Language}, pages 81--138.

\bibitem[{Dyer et~al.(2016)Dyer, Kuncoro, Ballesteros, and
  Smith}]{dyer-etal-2016-recurrent}
Dyer, Chris, Adhiguna Kuncoro, Miguel Ballesteros, and Noah~A. Smith. 2016.
\newblock Recurrent neural network grammars.
\newblock In \emph{Proceedings of the 2016 Conference of the North {A}merican
  Chapter of the Association for Computational Linguistics: Human Language
  Technologies}, pages 199--209, Association for Computational Linguistics, San
  Diego, California.

\bibitem[{Evans and Levinson(2009)}]{evans2009myth}
Evans, Nicholas and Stephen~C Levinson. 2009.
\newblock The myth of language universals: Language diversity and its
  importance for cognitive science.
\newblock \emph{Behavioral and brain sciences}, 32(5):429--448.

\bibitem[{Futrell, Levy, and Gibson(2020)}]{futrell2020dependency}
Futrell, Richard, Roger~P Levy, and Edward Gibson. 2020.
\newblock Dependency locality as an explanatory principle for word order.
\newblock \emph{Language}, 96(2):371--412.

\bibitem[{Futrell et~al.(2019)Futrell, Wilcox, Morita, Qian, Ballesteros, and
  Levy}]{futrell2019neural}
Futrell, Richard, Ethan Wilcox, Takashi Morita, Peng Qian, Miguel Ballesteros,
  and Roger Levy. 2019.
\newblock Neural language models as psycholinguistic subjects: Representations
  of syntactic state.
\newblock In \emph{Proceedings of the 2019 Conference of the North American
  Chapter of the Association for Computational Linguistics: Human Language
  Technologies, Volume 1 (Long and Short Papers)}, pages 32--42.

\bibitem[{Gauthier et~al.(2020)Gauthier, Hu, Wilcox, Qian, and
  Levy}]{gauthier2020syntaxgym}
Gauthier, Jon, Jennifer Hu, Ethan Wilcox, Peng Qian, and Roger Levy. 2020.
\newblock Syntaxgym: An online platform for targeted evaluation of language
  models.
\newblock In \emph{Proceedings of the 58th Annual Meeting of the Association
  for Computational Linguistics: System Demonstrations}, pages 70--76.

\bibitem[{Gerdes et~al.(2018)Gerdes, Guillaume, Kahane, and
  Perrier}]{gerdes2018sud}
Gerdes, Kim, Bruno Guillaume, Sylvain Kahane, and Guy Perrier. 2018.
\newblock Sud or surface-syntactic universal dependencies: An annotation scheme
  near-isomorphic to ud.
\newblock \emph{EMNLP 2018}, page~66.

\bibitem[{Gibson(1998)}]{gibson1998linguistic}
Gibson, Edward. 1998.
\newblock Linguistic complexity: Locality of syntactic dependencies.
\newblock \emph{Cognition}, 68(1):1--76.

\bibitem[{Gibson et~al.(2019)Gibson, Futrell, Piantadosi, Dautriche, Mahowald,
  Bergen, and Levy}]{gibson2019efficiency}
Gibson, Edward, Richard Futrell, Steven~P Piantadosi, Isabelle Dautriche, Kyle
  Mahowald, Leon Bergen, and Roger Levy. 2019.
\newblock How efficiency shapes human language.
\newblock \emph{Trends in cognitive sciences}, 23(5):389--407.

\bibitem[{Gibson et~al.(2000)}]{gibson2000dependency}
Gibson, Edward et~al. 2000.
\newblock The dependency locality theory: A distance-based theory of linguistic
  complexity.
\newblock \emph{Image, language, brain}, 2000:95--126.

\bibitem[{Giv{\'o}n(1995)}]{givon1995functionalism}
Giv{\'o}n, Talmy. 1995.
\newblock \emph{Functionalism and grammar}.
\newblock John Benjamins Publishing.

\bibitem[{Glava{\v{s}} and
  Vuli{\'c}(2021{\natexlab{a}})}]{glavavs2021supervised}
Glava{\v{s}}, Goran and Ivan Vuli{\'c}. 2021{\natexlab{a}}.
\newblock Is supervised syntactic parsing beneficial for language understanding
  tasks? an empirical investigation.
\newblock In \emph{Proceedings of the 16th Conference of the European Chapter
  of the Association for Computational Linguistics: Main Volume}, pages
  3090--3104.

\bibitem[{Glava{\v{s}} and Vuli{\'c}(2021{\natexlab{b}})}]{glavas21}
Glava{\v{s}}, Goran and Ivan Vuli{\'c}. 2021{\natexlab{b}}.
\newblock Is supervised syntactic parsing beneficial for language understanding
  tasks? an empirical investigation.
\newblock In \emph{Proceedings of the 16th Conference of the European Chapter
  of the Association for Computational Linguistics: Main Volume}, pages
  3090--3104.

\bibitem[{Gulordava et~al.(2018)Gulordava, Bojanowski, Grave, Linzen, and
  Baroni}]{gulordava2018colorless}
Gulordava, Kristina, Piotr Bojanowski, {\'E}douard Grave, Tal Linzen, and Marco
  Baroni. 2018.
\newblock Colorless green recurrent networks dream hierarchically.
\newblock In \emph{Proceedings of the 2018 Conference of the North American
  Chapter of the Association for Computational Linguistics: Human Language
  Technologies, Volume 1 (Long Papers)}, pages 1195--1205.

\bibitem[{Gupta, Kvernadze, and Srikumar(2021)}]{gupta2021bert}
Gupta, Ashim, Giorgi Kvernadze, and Vivek Srikumar. 2021.
\newblock Bert \& family eat word salad: Experiments with text understanding.
\newblock In \emph{Proceedings of the AAAI Conference on Artificial
  Intelligence}, volume~35, pages 12946--12954.

\bibitem[{Gururangan et~al.(2018)Gururangan, Swayamdipta, Levy, Schwartz,
  Bowman, and Smith}]{gururangan-etal-2018-annotation}
Gururangan, Suchin, Swabha Swayamdipta, Omer Levy, Roy Schwartz, Samuel Bowman,
  and Noah~A. Smith. 2018.
\newblock Annotation artifacts in natural language inference data.
\newblock In \emph{Proceedings of the 2018 Conference of the North {A}merican
  Chapter of the Association for Computational Linguistics: Human Language
  Technologies, Volume 2 (Short Papers)}, pages 107--112, Association for
  Computational Linguistics, New Orleans, Louisiana.

\bibitem[{Hahn, Jurafsky, and Futrell(2020)}]{hahn2020universals}
Hahn, Michael, Dan Jurafsky, and Richard Futrell. 2020.
\newblock Universals of word order reflect optimization of grammars for
  efficient communication.
\newblock \emph{Proceedings of the National Academy of Sciences},
  117(5):2347--2353.

\bibitem[{Hawkins(2004)}]{hawkins2004efficiency}
Hawkins, John~A. 2004.
\newblock \emph{Efficiency and complexity in grammars}.
\newblock Oxford University Press on Demand.

\bibitem[{Hewitt and Liang(2019)}]{hewitt-liang-2019-designing}
Hewitt, John and Percy Liang. 2019.
\newblock Designing and interpreting probes with control tasks.
\newblock In \emph{Proceedings of the 2019 Conference on Empirical Methods in
  Natural Language Processing and the 9th International Joint Conference on
  Natural Language Processing (EMNLP-IJCNLP)}, pages 2733--2743, Association
  for Computational Linguistics, Hong Kong, China.

\bibitem[{Hewitt and Manning(2019)}]{hewitt19}
Hewitt, John and Christopher~D. Manning. 2019.
\newblock A structural probe for finding syntax in word representations.

\bibitem[{Hochreiter and Schmidhuber(1997)}]{hochreiter97}
Hochreiter, Sepp and J{\"u}rgen Schmidhuber. 1997.
\newblock Long short-term memory.
\newblock \emph{Neural Computation}, 9(8):1735--1780.

\bibitem[{Howard and Ruder(2018)}]{howard-ruder-2018-universal}
Howard, Jeremy and Sebastian Ruder. 2018.
\newblock Universal language model fine-tuning for text classification.
\newblock In \emph{Proceedings of the 56th Annual Meeting of the Association
  for Computational Linguistics (Volume 1: Long Papers)}, pages 328--339,
  Association for Computational Linguistics, Melbourne, Australia.

\bibitem[{Htut et~al.(2019)Htut, Phang, Bordia, and Bowman}]{htut2019attention}
Htut, Phu~Mon, Jason Phang, Shikha Bordia, and Samuel~R Bowman. 2019.
\newblock Do attention heads in bert track syntactic dependencies?
\newblock \emph{arXiv preprint arXiv:1911.12246}.

\bibitem[{Hu et~al.(2020)Hu, Gauthier, Qian, Wilcox, and
  Levy}]{hu2020systematic}
Hu, Jennifer, Jon Gauthier, Peng Qian, Ethan Wilcox, and Roger~P Levy. 2020.
\newblock A systematic assessment of syntactic generalization in neural
  language models.
\newblock \emph{arXiv preprint arXiv:2005.03692}.

\bibitem[{Hupkes, Veldhoen, and Zuidema(2018)}]{hupkes18}
Hupkes, Dieuwke, Sara Veldhoen, and Willem Zuidema. 2018.
\newblock Visualization and `diagnostic classifiers' reveal how recurrent and
  recursive neural networks process hierarchical structure.
\newblock \emph{Journal of Artificial Intelligence Research}, 61:907--926.

\bibitem[{Ivanova et~al.(2012)Ivanova, Oepen, {\O}vrelid, and
  Flickinger}]{ivanova2012did}
Ivanova, Angelina, Stephan Oepen, Lilja {\O}vrelid, and Dan Flickinger. 2012.
\newblock Who did what to whom? a contrastive study of syntacto-semantic
  dependencies.
\newblock In \emph{Proceedings of the sixth linguistic annotation workshop},
  pages 2--11.

\bibitem[{Jacovi and Goldberg(2021)}]{jacovi2021aligning}
Jacovi, Alon and Yoav Goldberg. 2021.
\newblock Aligning faithful interpretations with their social attribution.
\newblock \emph{Transactions of the Association for Computational Linguistics},
  9:294--310.

\bibitem[{Jaeger and Tily(2011)}]{jaeger2011language}
Jaeger, T~Florian and Harry Tily. 2011.
\newblock On language ‘utility’: Processing complexity and communicative
  efficiency.
\newblock \emph{Wiley Interdisciplinary Reviews: Cognitive Science},
  2(3):323--335.

\bibitem[{Jain and Wallace(2019)}]{jain-wallace-2019-attention}
Jain, Sarthak and Byron~C. Wallace. 2019.
\newblock {A}ttention is not {E}xplanation.
\newblock In \emph{Proceedings of the 2019 Conference of the North {A}merican
  Chapter of the Association for Computational Linguistics: Human Language
  Technologies, Volume 1 (Long and Short Papers)}, pages 3543--3556,
  Association for Computational Linguistics, Minneapolis, Minnesota.

\bibitem[{Jumelet and Hupkes(2018)}]{jumelet-hupkes-2018-language}
Jumelet, Jaap and Dieuwke Hupkes. 2018.
\newblock Do language models understand anything? on the ability of {LSTM}s to
  understand negative polarity items.
\newblock In \emph{Proceedings of the 2018 {EMNLP} Workshop {B}lackbox{NLP}:
  Analyzing and Interpreting Neural Networks for {NLP}}, pages 222--231,
  Association for Computational Linguistics, Brussels, Belgium.

\bibitem[{Kulmizev et~al.(2020)Kulmizev, Ravishankar, Abdou, and
  Nivre}]{kulmizev20}
Kulmizev, Artur, Vinit Ravishankar, Mostafa Abdou, and Joakim Nivre. 2020.
\newblock Do neural language models show preferences for syntactic formalisms?
\newblock pages ?--?

\bibitem[{Kuncoro et~al.(2020)Kuncoro, Kong, Fried, Yogatama, Rimell, Dyer, and
  Blunsom}]{kuncoro2020syntactic}
Kuncoro, Adhiguna, Lingpeng Kong, Daniel Fried, Dani Yogatama, Laura Rimell,
  Chris Dyer, and Phil Blunsom. 2020.
\newblock Syntactic structure distillation pretraining for bidirectional
  encoders.
\newblock \emph{Transactions of the Association for Computational Linguistics},
  8:776--794.

\bibitem[{Lasnik and Lidz(2017)}]{lasnik2017argument}
Lasnik, Howard and Jeffrey~L Lidz. 2017.
\newblock The argument from the poverty of the stimulus.
\newblock \emph{The Oxford handbook of universal grammar}, pages 221--248.

\bibitem[{Linzen and Baroni(2021)}]{linzen2021syntactic}
Linzen, Tal and Marco Baroni. 2021.
\newblock Syntactic structure from deep learning.
\newblock \emph{Annual Review of Linguistics}, 7:195--212.

\bibitem[{Linzen, Dupoux, and
  Goldberg(2016{\natexlab{a}})}]{linzen2016assessing}
Linzen, Tal, Emmanuel Dupoux, and Yoav Goldberg. 2016{\natexlab{a}}.
\newblock Assessing the ability of lstms to learn syntax-sensitive
  dependencies.
\newblock \emph{Transactions of the Association for Computational Linguistics},
  4:521--535.

\bibitem[{Linzen, Dupoux, and Goldberg(2016{\natexlab{b}})}]{linzen16}
Linzen, Tal, Emmanuel Dupoux, and Yoav Goldberg. 2016{\natexlab{b}}.
\newblock Assessing the ability of {LSTM}s to learn syntax-sensitive
  dependencies.
\newblock \emph{Transactions of the Association for Computational Linguistics},
  4:521--535.

\bibitem[{Lipton(2018)}]{lipton2018mythos}
Lipton, Zachary~C. 2018.
\newblock The mythos of model interpretability: In machine learning, the
  concept of interpretability is both important and slippery.
\newblock \emph{Queue}, 16(3):31--57.

\bibitem[{Liu et~al.(2019{\natexlab{a}})Liu, Gardner, Belinkov, Peters, and
  Smith}]{liu-etal-2019-linguistic}
Liu, Nelson~F., Matt Gardner, Yonatan Belinkov, Matthew~E. Peters, and Noah~A.
  Smith. 2019{\natexlab{a}}.
\newblock Linguistic knowledge and transferability of contextual
  representations.
\newblock In \emph{Proceedings of the 2019 Conference of the North {A}merican
  Chapter of the Association for Computational Linguistics: Human Language
  Technologies, Volume 1 (Long and Short Papers)}, pages 1073--1094,
  Association for Computational Linguistics, Minneapolis, Minnesota.

\bibitem[{Liu et~al.(2019{\natexlab{b}})Liu, Ott, Goyal, Du, Joshi, Chen, Levy,
  Lewis, Zettlemoyer, and Stoyanov}]{liu2019roberta}
Liu, Yinhan, Myle Ott, Naman Goyal, Jingfei Du, Mandar Joshi, Danqi Chen, Omer
  Levy, Mike Lewis, Luke Zettlemoyer, and Veselin Stoyanov. 2019{\natexlab{b}}.
\newblock Roberta: A robustly optimized bert pretraining approach.
\newblock \emph{arXiv preprint arXiv:1907.11692}.

\bibitem[{Lundberg and Lee(2017)}]{lundberg2017unified}
Lundberg, Scott~M and Su-In Lee. 2017.
\newblock A unified approach to interpreting model predictions.
\newblock In \emph{Proceedings of the 31st international conference on neural
  information processing systems}, pages 4768--4777.

\bibitem[{Manning et~al.(2020)Manning, Clark, Hewitt, Khandelwal, and
  Levy}]{manning2020emergent}
Manning, Christopher~D, Kevin Clark, John Hewitt, Urvashi Khandelwal, and Omer
  Levy. 2020.
\newblock Emergent linguistic structure in artificial neural networks trained
  by self-supervision.
\newblock \emph{Proceedings of the National Academy of Sciences},
  117(48):30046--30054.

\bibitem[{de~Marneffe et~al.(2021)de~Marneffe, Manning, Nivre, and
  Zeman}]{demarneffe21}
de~Marneffe, Marie, Christopher~D. Manning, Joakim Nivre, and Daniel Zeman.
  2021.
\newblock Universal dependencies.
\newblock \emph{Computational Linguistics}, 47:255--308.

\bibitem[{de~Marneffe, MacCartney, and
  Manning(2006)}]{de-marneffe-etal-2006-generating}
de~Marneffe, Marie-Catherine, Bill MacCartney, and Christopher~D. Manning.
  2006.
\newblock Generating typed dependency parses from phrase structure parses.
\newblock In \emph{Proceedings of the Fifth International Conference on
  Language Resources and Evaluation ({LREC}{'}06)}, European Language Resources
  Association (ELRA), Genoa, Italy.

\bibitem[{Marvin and Linzen(2018)}]{marvin2018targeted}
Marvin, Rebecca and Tal Linzen. 2018.
\newblock Targeted syntactic evaluation of language models.
\newblock In \emph{Proceedings of the 2018 Conference on Empirical Methods in
  Natural Language Processing}, pages 1192--1202.

\bibitem[{Matthews(1981)}]{matthews81}
Matthews, Peter~H. 1981.
\newblock \emph{Syntax}.

\bibitem[{Maudslay et~al.(2020)Maudslay, Valvoda, Pimentel, Williams, and
  Cotterell}]{maudslay2020tale}
Maudslay, Rowan~Hall, Josef Valvoda, Tiago Pimentel, Adina Williams, and Ryan
  Cotterell. 2020.
\newblock A tale of a probe and a parser.
\newblock In \emph{Proceedings of the 58th Annual Meeting of the Association
  for Computational Linguistics}, pages 7389--7395.

\bibitem[{McCoy, Pavlick, and Linzen(2019)}]{mccoy-etal-2019-right}
McCoy, Tom, Ellie Pavlick, and Tal Linzen. 2019.
\newblock Right for the wrong reasons: Diagnosing syntactic heuristics in
  natural language inference.
\newblock In \emph{Proceedings of the 57th Annual Meeting of the Association
  for Computational Linguistics}, pages 3428--3448, Association for
  Computational Linguistics, Florence, Italy.

\bibitem[{Mel'{\v{c}}uk(1988)}]{melcuk88}
Mel'{\v{c}}uk, Igor. 1988.
\newblock \emph{Dependency Syntax: Theory and Practice}.
\newblock State University of New York Press.

\bibitem[{Mikolov et~al.(2013)Mikolov, Chen, Corrado, and Dean}]{mikolov13}
Mikolov, Tomas, Kai Chen, Greg Corrado, and Jeffrey Dean. 2013.
\newblock Efficient estimation of word representations in vector space.
\newblock \emph{arXiv preprint arXiv:1301.3781}.

\bibitem[{Mollica et~al.(2020)Mollica, Siegelman, Diachek, Piantadosi,
  Mineroff, Futrell, Kean, Qian, and Fedorenko}]{mollica2020composition}
Mollica, Francis, Matthew Siegelman, Evgeniia Diachek, Steven~T Piantadosi,
  Zachary Mineroff, Richard Futrell, Hope Kean, Peng Qian, and Evelina
  Fedorenko. 2020.
\newblock Composition is the core driver of the language-selective network.
\newblock \emph{Neurobiology of Language}, 1(1):104--134.

\bibitem[{Newman et~al.(2021)Newman, Ang, Gong, and Hewitt}]{newman21}
Newman, Benjamin, Kai{-}Siang Ang, Julia Gong, and John Hewitt. 2021.
\newblock Refining targeted syntactic evaluation of language models.
\newblock \emph{CoRR}, abs/2104.09635.

\bibitem[{Pennington, Socher, and Manning(2014)}]{pennington14}
Pennington, Jeffrey, Richard Socher, and Christopher Manning. 2014.
\newblock {G}love: Global vectors for word representation.
\newblock pages 1532--1543.

\bibitem[{Peters et~al.(2018)Peters, Neumann, Iyyer, Gardner, clark, Lee, and
  Zettlemoyer}]{peters18}
Peters, Matthew~E., Mark Neumann, Mohit Iyyer, Matt Gardner, Christopher clark,
  Kenton Lee, and Luke Zettlemoyer. 2018.
\newblock Deep contextualized word representations.
\newblock pages 2227--2237.

\bibitem[{Pham et~al.(2020)Pham, Bui, Mai, and Nguyen}]{pham2020out}
Pham, Thang~M, Trung Bui, Long Mai, and Anh Nguyen. 2020.
\newblock Out of order: How important is the sequential order of words in a
  sentence in natural language understanding tasks?
\newblock \emph{arXiv preprint arXiv:2012.15180}.

\bibitem[{Pimentel et~al.(2020)Pimentel, Saphra, Williams, and
  Cotterell}]{pimentel-etal-2020-pareto}
Pimentel, Tiago, Naomi Saphra, Adina Williams, and Ryan Cotterell. 2020.
\newblock {P}areto probing: {T}rading off accuracy for complexity.
\newblock In \emph{Proceedings of the 2020 Conference on Empirical Methods in
  Natural Language Processing (EMNLP)}, pages 3138--3153, Association for
  Computational Linguistics, Online.

\bibitem[{Poerner, Sch{\"u}tze, and Roth(2018)}]{poerner-etal-2018-evaluating}
Poerner, Nina, Hinrich Sch{\"u}tze, and Benjamin Roth. 2018.
\newblock Evaluating neural network explanation methods using hybrid documents
  and morphosyntactic agreement.
\newblock In \emph{Proceedings of the 56th Annual Meeting of the Association
  for Computational Linguistics (Volume 1: Long Papers)}, pages 340--350,
  Association for Computational Linguistics, Melbourne, Australia.

\bibitem[{Poliak et~al.(2018)Poliak, Naradowsky, Haldar, Rudinger, and
  Van~Durme}]{poliak-etal-2018-hypothesis}
Poliak, Adam, Jason Naradowsky, Aparajita Haldar, Rachel Rudinger, and Benjamin
  Van~Durme. 2018.
\newblock Hypothesis only baselines in natural language inference.
\newblock In \emph{Proceedings of the Seventh Joint Conference on Lexical and
  Computational Semantics}, pages 180--191, Association for Computational
  Linguistics, New Orleans, Louisiana.

\bibitem[{Radford et~al.(2019)Radford, Wu, Child, Luan, Amodei, Sutskever
  et~al.}]{radford2019language}
Radford, Alec, Jeffrey Wu, Rewon Child, David Luan, Dario Amodei, Ilya
  Sutskever, et~al. 2019.
\newblock Language models are unsupervised multitask learners.
\newblock \emph{OpenAI blog}, 1(8):9.

\bibitem[{Rajani et~al.(2019)Rajani, McCann, Xiong, and
  Socher}]{rajani-etal-2019-explain}
Rajani, Nazneen~Fatema, Bryan McCann, Caiming Xiong, and Richard Socher. 2019.
\newblock Explain yourself! leveraging language models for commonsense
  reasoning.
\newblock In \emph{Proceedings of the 57th Annual Meeting of the Association
  for Computational Linguistics}, pages 4932--4942, Association for
  Computational Linguistics, Florence, Italy.

\bibitem[{Ravfogel, Goldberg, and Tyers(2018)}]{ravfogel-etal-2018-lstm}
Ravfogel, Shauli, Yoav Goldberg, and Francis Tyers. 2018.
\newblock Can {LSTM} learn to capture agreement? the case of {B}asque.
\newblock In \emph{Proceedings of the 2018 {EMNLP} Workshop {B}lackbox{NLP}:
  Analyzing and Interpreting Neural Networks for {NLP}}, pages 98--107,
  Association for Computational Linguistics, Brussels, Belgium.

\bibitem[{Ribeiro, Singh, and Guestrin(2016)}]{ribeiro2016should}
Ribeiro, Marco~Tulio, Sameer Singh, and Carlos Guestrin. 2016.
\newblock " why should i trust you?" explaining the predictions of any
  classifier.
\newblock In \emph{Proceedings of the 22nd ACM SIGKDD international conference
  on knowledge discovery and data mining}, pages 1135--1144.

\bibitem[{Rimell, Clark, and Steedman(2009)}]{rimell09}
Rimell, Laura, Stephen Clark, and Mark Steedman. 2009.
\newblock Unbounded dependency recovery for parser evaluation.
\newblock In \emph{Proceedings of the 2009 Conference on Empirical Methods in
  Natural Language Processing}, pages 813--821.

\bibitem[{Schlichtkrull et~al.(2018)Schlichtkrull, Kipf, Bloem, Van Den~Berg,
  Titov, and Welling}]{schlichtkrull2018modeling}
Schlichtkrull, Michael, Thomas~N Kipf, Peter Bloem, Rianne Van Den~Berg, Ivan
  Titov, and Max Welling. 2018.
\newblock Modeling relational data with graph convolutional networks.
\newblock In \emph{European semantic web conference}, pages 593--607, Springer.

\bibitem[{Serrano and Smith(2019)}]{serrano-smith-2019-attention}
Serrano, Sofia and Noah~A. Smith. 2019.
\newblock Is attention interpretable?
\newblock In \emph{Proceedings of the 57th Annual Meeting of the Association
  for Computational Linguistics}, pages 2931--2951, Association for
  Computational Linguistics, Florence, Italy.

\bibitem[{Simonyan, Vedaldi, and Zisserman(2013)}]{simonyan2013deep}
Simonyan, Karen, Andrea Vedaldi, and Andrew Zisserman. 2013.
\newblock Deep inside convolutional networks: Visualising image classification
  models and saliency maps.
\newblock \emph{arXiv preprint arXiv:1312.6034}.

\bibitem[{Sinha et~al.(2021)Sinha, Jia, Hupkes, Pineau, Williams, and
  Kiela}]{sinha21}
Sinha, Koustuv, Robin Jia, Dieuwke Hupkes, Joelle Pineau, Adina Williams, and
  Douwe Kiela. 2021.
\newblock Masked language modeling and the distributional hypothesis: Order
  word matters pre-training for little.
\newblock \emph{CoRR}, abs/2104.06644.

\bibitem[{Sinha et~al.(2020)Sinha, Parthasarathi, Pineau, and
  Williams}]{sinha2020unnatural}
Sinha, Koustuv, Prasanna Parthasarathi, Joelle Pineau, and Adina Williams.
  2020.
\newblock Unnatural language inference.
\newblock \emph{arXiv preprint arXiv:2101.00010}.

\bibitem[{Sundararajan, Taly, and Yan(2017)}]{sundararajan2017axiomatic}
Sundararajan, Mukund, Ankur Taly, and Qiqi Yan. 2017.
\newblock Axiomatic attribution for deep networks.
\newblock In \emph{International Conference on Machine Learning}, pages
  3319--3328, PMLR.

\bibitem[{Swayamdipta et~al.(2019)Swayamdipta, Peters, Roof, Dyer, and
  Smith}]{swayamdipta2019shallow}
Swayamdipta, Swabha, Matthew Peters, Brendan Roof, Chris Dyer, and Noah~A
  Smith. 2019.
\newblock Shallow syntax in deep water.
\newblock \emph{arXiv preprint arXiv:1908.11047}.

\bibitem[{Tenney, Das, and Pavlick(2019)}]{tenney-etal-2019-bert}
Tenney, Ian, Dipanjan Das, and Ellie Pavlick. 2019.
\newblock {BERT} rediscovers the classical {NLP} pipeline.
\newblock In \emph{Proceedings of the 57th Annual Meeting of the Association
  for Computational Linguistics}, pages 4593--4601, Association for
  Computational Linguistics, Florence, Italy.

\bibitem[{Tenney et~al.(2019)Tenney, Xia, Chen, Wang, Poliak, McCoy, Kim,
  Van~Durme, Bowman, Das et~al.}]{tenney2019you}
Tenney, Ian, Patrick Xia, Berlin Chen, Alex Wang, Adam Poliak, R~Thomas McCoy,
  Najoung Kim, Benjamin Van~Durme, Samuel~R Bowman, Dipanjan Das, et~al. 2019.
\newblock What do you learn from context? probing for sentence structure in
  contextualized word representations.
\newblock \emph{arXiv preprint arXiv:1905.06316}.

\bibitem[{Tesni{\`e}re(1959)}]{tesniere59}
Tesni{\`e}re, Lucien. 1959.
\newblock \emph{{\'E}l{\'e}ments de syntaxe structurale}.
\newblock Editions Klincksieck.

\bibitem[{Tomasello(2009)}]{tomasello2009cultural}
Tomasello, Michael. 2009.
\newblock \emph{The cultural origins of human cognition}.
\newblock Harvard university press.

\bibitem[{Traxler(2014)}]{traxler2014trends}
Traxler, Matthew~J. 2014.
\newblock Trends in syntactic parsing: Anticipation, bayesian estimation, and
  good-enough parsing.
\newblock \emph{Trends in cognitive sciences}, 18(11):605--611.

\bibitem[{Trudgill(2017)}]{trudgill2017anthropological}
Trudgill, Peter. 2017.
\newblock The anthropological setting of polysynthesis.
\newblock In \emph{The Oxford handbook of polysynthesis}.

\bibitem[{Wang et~al.(2019)Wang, Pruksachatkun, Nangia, Singh, Michael, Hill,
  Levy, and Bowman}]{wang2019superglue}
Wang, Alex, Yada Pruksachatkun, Nikita Nangia, Amanpreet Singh, Julian Michael,
  Felix Hill, Omer Levy, and Samuel~R Bowman. 2019.
\newblock Superglue: A stickier benchmark for general-purpose language
  understanding systems.
\newblock \emph{arXiv preprint arXiv:1905.00537}.

\bibitem[{Wang et~al.(2018)Wang, Singh, Michael, Hill, Levy, and
  Bowman}]{wang2018glue}
Wang, Alex, Amanpreet Singh, Julian Michael, Felix Hill, Omer Levy, and
  Samuel~R Bowman. 2018.
\newblock Glue: A multi-task benchmark and analysis platform for natural
  language understanding.
\newblock \emph{arXiv preprint arXiv:1804.07461}.

\bibitem[{Warstadt et~al.(2020)Warstadt, Parrish, Liu, Mohananey, Peng, Wang,
  and Bowman}]{warstadt2020blimp}
Warstadt, Alex, Alicia Parrish, Haokun Liu, Anhad Mohananey, Wei Peng, Sheng-Fu
  Wang, and Samuel~R Bowman. 2020.
\newblock Blimp: The benchmark of linguistic minimal pairs for english.
\newblock \emph{Transactions of the Association for Computational Linguistics},
  8:377--392.

\bibitem[{Wiegreffe, Marasovi{\'c}, and Smith(2020)}]{wiegreffe2020measuring}
Wiegreffe, Sarah, Ana Marasovi{\'c}, and Noah~A Smith. 2020.
\newblock Measuring association between labels and free-text rationales.
\newblock \emph{arXiv preprint arXiv:2010.12762}.

\bibitem[{Wiegreffe and Pinter(2019)}]{wiegreffe-pinter-2019-attention}
Wiegreffe, Sarah and Yuval Pinter. 2019.
\newblock Attention is not not explanation.
\newblock In \emph{Proceedings of the 2019 Conference on Empirical Methods in
  Natural Language Processing and the 9th International Joint Conference on
  Natural Language Processing (EMNLP-IJCNLP)}, pages 11--20, Association for
  Computational Linguistics, Hong Kong, China.

\bibitem[{Wilcox et~al.(2018)Wilcox, Levy, Morita, and
  Futrell}]{wilcox-etal-2018-rnn}
Wilcox, Ethan, Roger Levy, Takashi Morita, and Richard Futrell. 2018.
\newblock What do {RNN} language models learn about filler{--}gap dependencies?
\newblock In \emph{Proceedings of the 2018 {EMNLP} Workshop {B}lackbox{NLP}:
  Analyzing and Interpreting Neural Networks for {NLP}}, pages 211--221,
  Association for Computational Linguistics, Brussels, Belgium.

\bibitem[{Wu, Peng, and Smith(2021)}]{wu2021infusing}
Wu, Zhaofeng, Hao Peng, and Noah~A Smith. 2021.
\newblock Infusing finetuning with semantic dependencies.
\newblock \emph{Transactions of the Association for Computational Linguistics},
  9:226--242.

\bibitem[{Zipf(1949)}]{zipf1949human}
Zipf, G.K. 1949.
\newblock \emph{Human Behavior and the Principle of Least Effort: An
  Introduction to Human Ecology}.
\newblock Addison-Wesley Press.

\end{thebibliography}

\end{document}